\pdfoutput=1

\documentclass[11pt]{article}

\usepackage[]{ACL2023}

\usepackage{times}
\usepackage{latexsym}
\usepackage{amsmath}
\usepackage{multirow}
\usepackage{graphicx}
\usepackage{amsfonts}

\usepackage[T1]{fontenc}

\usepackage[utf8]{inputenc}
\usepackage[figuresright]{rotating}
\usepackage{microtype}

\usepackage{inconsolata}
\usepackage{booktabs}
\usepackage{amsfonts}
\title{\textsc{SHINE}: Syntax-augmented Hierarchical Interactive Encoder for \\ Zero-shot Cross-lingual Information Extraction}

\author{
Jun-Yu Ma$^1$, Jia-Chen Gu$^1$, Zhen-Hua Ling$^1$, 
{\bf Quan Liu$^{2,3}$, Cong Liu$^{1,3}$, Guoping Hu$^{2,3}$} \\
  $^1$National Engineering Research Center of Speech and Language Information Processing, \\
      University of Science and Technology of China, Hefei, China \\
  $^2$State Key Laboratory of Cognitive Intelligence ~ 
  $^3$iFLYTEK Research, Hefei, China \\
{\tt \{mjy1999\}@mail.ustc.edu.cn}, {\tt \{gujc,zhling\}@ustc.edu.cn}, \\ {\tt \{quanliu,congliu2,gphu\}@iflytek.com}
}

\begin{document}
\maketitle
\begin{abstract}
Zero-shot cross-lingual information extraction
(IE) aims at constructing an IE model
for some low-resource target languages, given annotations exclusively in some rich-resource languages.
Recent studies based on language-universal features have shown their effectiveness and are attracting increasing attention.
However, prior work has neither explored the potential of establishing interactions between language-universal features and contextual representations nor incorporated features that can effectively model constituent span attributes and relationships between multiple spans.
In this study, a \textbf{s}yntax-augmented \textbf{h}ierarchical \textbf{in}teractive \textbf{e}ncoder (SHINE) is proposed to transfer cross-lingual IE knowledge. 
The proposed encoder is capable of interactively capturing complementary information between features and contextual information, to derive language-agnostic representations for various IE tasks.
Concretely, a multi-level interaction network is designed to hierarchically interact the complementary information to strengthen domain adaptability.
Besides, in addition to the well-studied syntax features of part-of-speech and dependency relation, a new syntax feature of constituency structure is introduced to model the constituent span information which is crucial for IE.
Experiments across seven languages on three IE tasks and four benchmarks verify the effectiveness and generalization ability of the proposed method.
\end{abstract}

\section{Introduction}

Information Extraction (IE) aims at extracting structured information from unstructured texts to classify and reconstruct massive amounts of content automatically.
It covers a great variety of tasks, such as named entity recognition (NER)~\cite{DBLP:conf/acl/LiFMHWL20,DBLP:journals/corr/abs-2305-02517}, relation extraction~\cite{DBLP:conf/acl/AbadNM17,DBLP:conf/emnlp/BekoulisDDD18} and event extraction~\cite{DBLP:conf/acl/ZhouGH16,DBLP:conf/acl/ChenLZLZ17}.
Recently, thanks to the breakthroughs of deep learning, neural network models have achieved significant improvements in various IE tasks~\cite{DBLP:conf/naacl/LuanWHSOH19,DBLP:conf/acl/LinJHW20,DBLP:conf/semeval/ChenMQGLL22}.
However, it is too expensive to annotate a large amount of data in low-resource languages for supervised IE training.
Therefore, cross-lingual IE under low-resource~\cite{DBLP:conf/propor/CabralGSC20,DBLP:conf/emnlp/YarmohammadiWMX21,DBLP:conf/sigir/Agirre22} settings has attracted considerable attention. In this paper, we study zero-shot cross-lingual IE, where annotated training data is available only in the source language, but not in the target language.

\begin{figure}[t]
\centering
\includegraphics[width=0.45\textwidth]{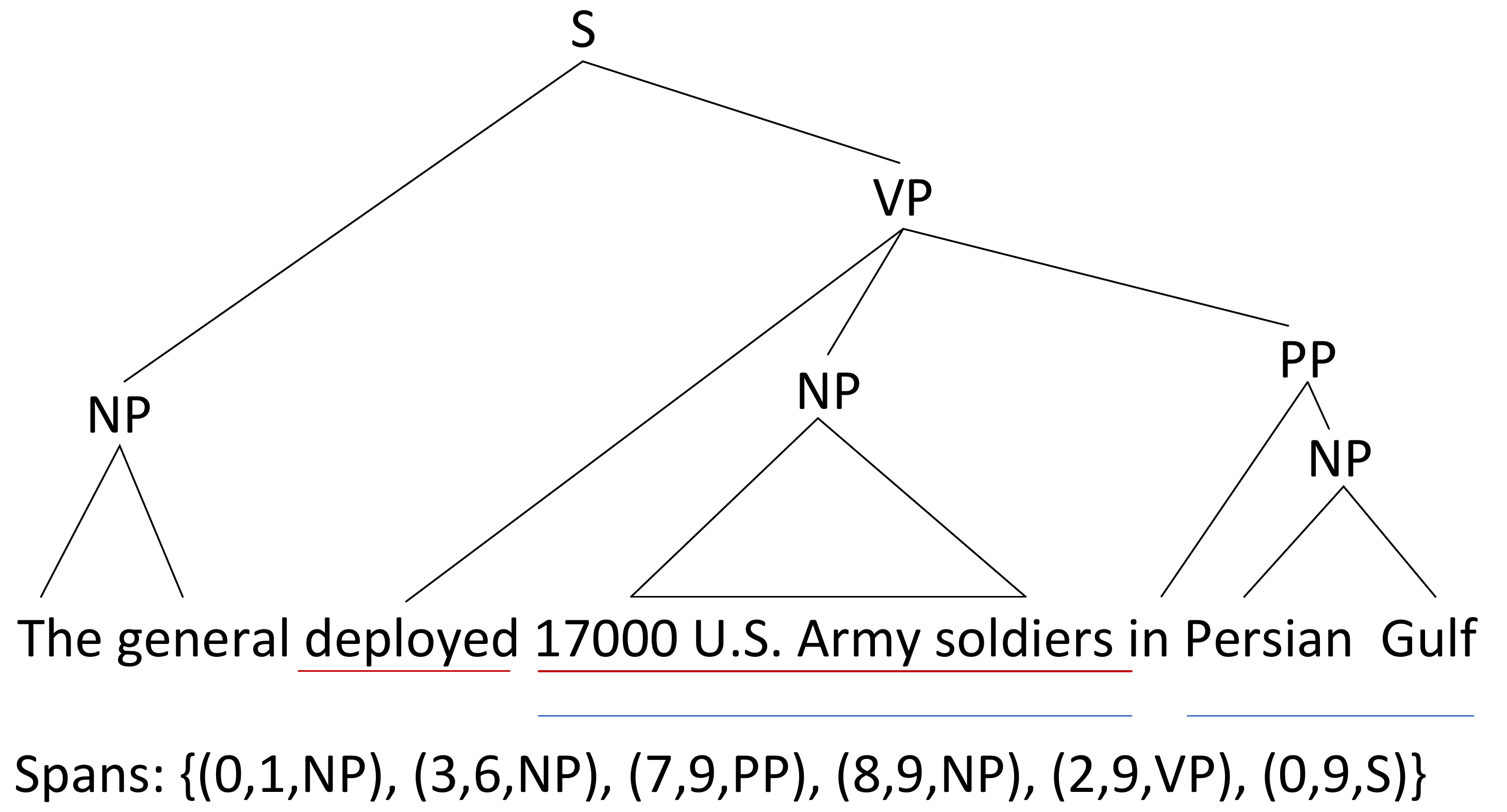}
\caption{An example of constituency tree, including: 
(1) an event trigger (\emph{\textcolor{red}{deployed}}) and its argument (\emph{\textcolor{red}{17000 U.S. Army soldiers}}), and
(2) two entities corresponding to subject (\emph{\textcolor{blue}{17000 U.S. Army soldiers}}) and object (\emph{\textcolor{blue}{Persian Gulf}}) respectively.
}  
\label{fig1}
\vspace{-6mm}
\end{figure}

Existing methods on reducing the need for annotated data in cross-lingual IE tasks can be generally categorized into shared representation space-based~\cite{DBLP:conf/conll/TsaiMR16,DBLP:conf/emnlp/WuD19,DBLP:conf/conll/MhamdiFM19}, translation-based~\cite{DBLP:conf/emnlp/MayhewTR17,DBLP:conf/emnlp/YarmohammadiWMX21,DBLP:conf/sigir/LouGYWZTX22} and language-universal features-based methods~\cite{DBLP:conf/acl/DaganJVHCR18,DBLP:conf/emnlp/SubburathinamLJ19,DBLP:conf/aaai/AhmadPC21}. 
In this work, we study the last one.
\citet{DBLP:conf/emnlp/SubburathinamLJ19} have verified that language-universal symbolic and distributional representations are complementary for cross-lingual structure transfer.
Language-universal features-based methods aggregate universal and complementary information 
to obtain schema consistency across languages for effective knowledge transfer~\cite{DBLP:conf/emnlp/LiuCLZ19,DBLP:conf/emnlp/SubburathinamLJ19,DBLP:conf/aaai/AhmadPC21}. 
For example, \citet{DBLP:conf/aaai/AhmadPC21} utilized part-of-speech (POS), entity type and dependency features to construct a cross-lingual transfer framework. 


Although language-universal features-based methods have achieved competitive performance on zero-shot cross-lingual IE, the features used in previous work are still incapable of capturing sufficient contextualized semantics. 
Existing methods have never explored the potential of establishing interactions between these features and contextual representations, which leads to a representation gap between them and information loss. 
Besides, previous studies focus on employing dependency structures for various IE tasks~\cite{DBLP:conf/emnlp/SubburathinamLJ19,DBLP:conf/aaai/AhmadPC21}, describing only the dependency relationships between two words.
Essentially, most IE tasks aim at identifying and categorizing phrase spans, thus the span-level information such as the constituent span attributes and the relationships between multiple spans are crucial. 
However, capturing this information is well beyond the scope of features studied extensively in prior work~\cite{DBLP:conf/aaai/AhmadPC21}.
Here, we give an example to illustrate the importance of constituent span information as shown in Figure~\ref{fig1}.
For event extraction, given an event trigger \textit{deployed} and two argument candidates \textit{U.S. Army} and \textit{17000 U.S. Army soldiers}, the former candidate may be recognized as an argument without any guidance.
However, the latter candidate is a constituent span which shares the same parent node labeled VP with \textit{deployed}, then it can be recognized correctly as an argument under this signal.
This property is common in many languages, so it is beneficial to be able to model this universal information when transferred across languages.


On account of the above issues, a \textbf{s}yntax-augmented \textbf{h}ierarchical \textbf{in}teractive \textbf{e}ncoder (SHINE)
is proposed in this paper to transfer cross-lingual IE knowledge.
It is capable of interactively capturing complementary information between language-universal features and contextual information, so as to derive language-agnostic contextualized representations for various IE tasks.
On the one hand, a multi-level interaction network is designed to encourage the distributions of features and contextual representations to approximate each other at three levels via an interactive loss mechanism. 
Specifically, the global-level interaction operates on the entire sentence,
the local-level one operates on sub-spans in a sentence, 
and the task-level one operates on task-related mentions.
In this way, the model can hierarchically interact the  complementary information to strengthen its domain adaptability.
Features of POS, dependency relation and entity type used in previous studies are adopted to verify the effectiveness of the proposed method.\footnote{Although entity type is not formally a syntax feature, it is still adopted in this work following \citet{DBLP:conf/aaai/AhmadPC21}.}
Additionally, a new syntax feature of constituency structure is introduced to explicitly utilize the span-level information in the text. 
Considering the overlap between constituent spans, these spans are first converted into  word-level embeddings.
Then a frequency matrix is designed where each element represents the number of occurrences of a sub-span in all constituent spans.
These not only enrich the attribute information of constituent spans,
but also model the importance of each sub-span,
so that task-related spans can be accurately captured for effective cross-language transfer.


To measure the effectiveness of the proposed method and to test its generalization ability, SHINE is evaluated on three IE tasks including NER, relation extraction and event extraction.
Experiments across seven languages on four benchmarks are conducted for evaluation.
Results show that SHINE achieves highly promising performance, verifying the importance of interactions between universal features and contextual representations, as well as the effectiveness of constituency structure for IE.
To facilitate others to reproduce our results, we will publish all source code later.


In summary, our contributions in this paper are three-fold:
(1) A multi-level interaction framework is proposed to
interact the complementary structured information contained in language-universal features and contextual representations.
(2) The constituency feature is first introduced to explicitly utilize the constituent span information for cross-lingual IE.
(3) Experiments across seven languages on three tasks and four benchmarks verify the effectiveness and generalization ability of SHINE.



\section{Related Work}
Many researchers have investigated shared representation space-based, translation-based and language-universal features-based methods for zero-shot cross-lingual IE tasks.
Shared representation space-based models capture features of labeled source-language data using multilingual pre-trained models, and then they are applied to the target languages directly~\cite{DBLP:conf/conll/TsaiMR16,DBLP:conf/emnlp/WuD19,DBLP:conf/conll/MhamdiFM19}.
However, this type of methods can transfer only superficial information due to representation discrepancy between source and target languages.
Besides, translation-based methods translate texts from the source language to the target languages, and then project annotations accordingly to create silver annotations~\cite{DBLP:conf/emnlp/MayhewTR17,DBLP:conf/sigir/LouGYWZTX22}. But noise from translation and projection might degrade performance.
Language-universal features-based methods are effective in cross-lingual IE by utilizing universal and complementary information 
to learn multi-lingual common space representations~\cite{DBLP:conf/emnlp/LiuCLZ19,DBLP:conf/emnlp/SubburathinamLJ19,DBLP:conf/aaai/AhmadPC21}.
~\citet{DBLP:conf/emnlp/LiuCLZ19} utilized GCN~\cite{DBLP:conf/iclr/KipfW17} 
to learn representations based on universal dependency parses
to improve cross-lingual transfer for IE.
\citet{DBLP:conf/emnlp/SubburathinamLJ19} exploited other features such as POS and entity type for embedding.
\citet{DBLP:conf/aaai/AhmadPC21} employed transformer~\cite{DBLP:conf/nips/VaswaniSPUJGKP17} to fuse structural information to learn the dependencies between words with different syntactic distances.

Compared with \citet{DBLP:conf/emnlp/SubburathinamLJ19} and \citet{DBLP:conf/aaai/AhmadPC21} that are the most relevant to this work, the main differences are highlighted.
These methods have never explored the potential of establishing interactions between language-universal features and contextual representations.
Besides, the well-studied features 
focus on word attributes and the relationship between words.
They cannot model span-level information such as the
constituent span attributes and the relationships between multiple spans which are crucial for cross-lingual IE.
To the best of our knowledge, 
this paper makes the first
attempt to interactively capture complementary information between universal features and contextual information, and to introduce constituency structure to explicitly utilize span-level information. 

\section{Methodology}
In this section, we present the detailed framework of the proposed SHINE.
For one thing, a multi-level interaction network is introduced. 
Specifically, three classes of adaptation methods are adopted to interact language-universal with contextual information via an
interactive loss mechanism.
In order to verify the effectiveness of the proposed framework, features such as POS, dependency relation and entity type used in previous studies are adopted.
Furthermore, a new syntax feature of constituency
structure is introduced to make up for the deficiencies of the existing work in explicitly modeling and utilizing the constituent span information.

\begin{figure*}[h]
\centering
\includegraphics[width=1.0\textwidth]{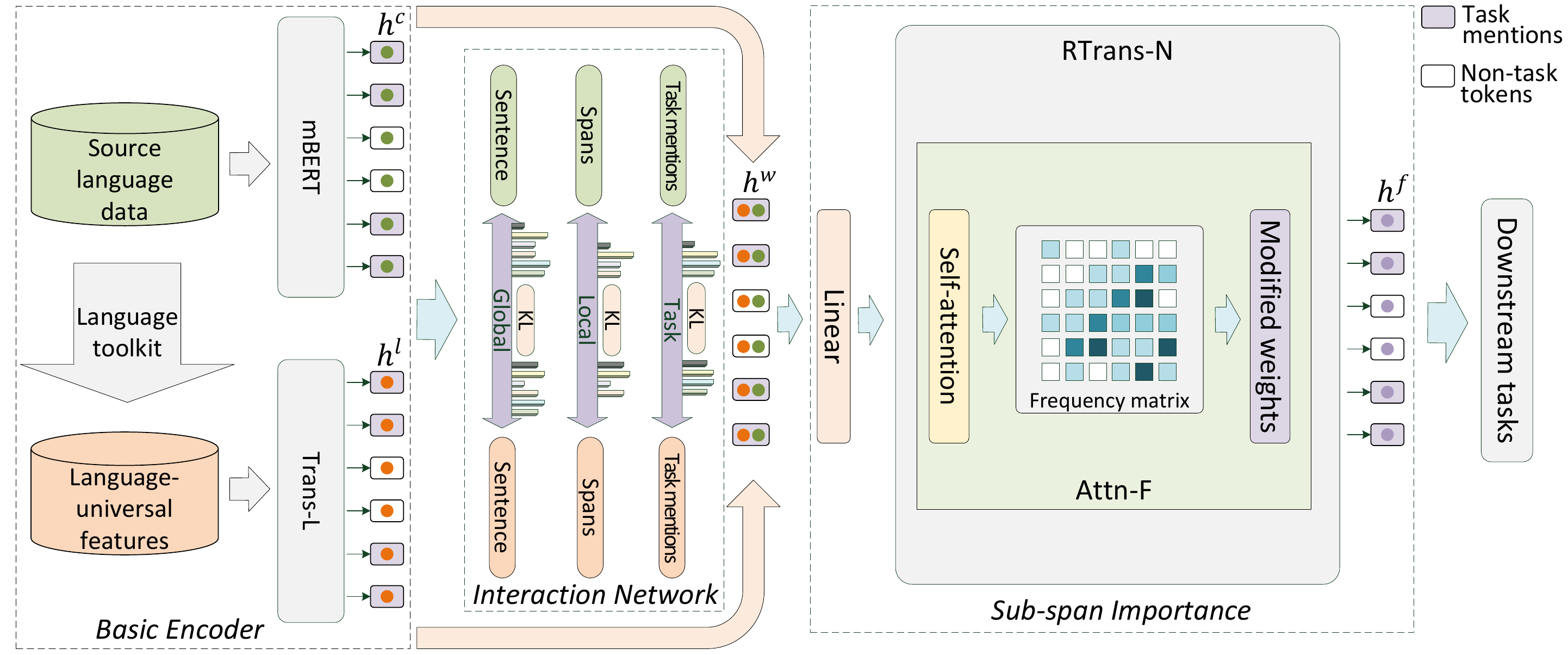}
\caption{The overall structure of the proposed SHINE.
$\boldsymbol{h^{c}}$ and $\boldsymbol{h^{l}}$ refer to the contextual and language-universal representations, respectively.
$\boldsymbol{h^{w}}$ is the concatenation of $\boldsymbol{h^{c}}$ and $\boldsymbol{h^{l}}$. 
$\boldsymbol{h^{f}}$ is the final representation of the model.
}
\label{modelshine}

\end{figure*}

\subsection{Problem Definition}

Denote one sentence as \(\boldsymbol{x}=\{x_i\}_{i=1}^{L}\) with its language-universal features \(\boldsymbol{x^l}\) and annotations \(\boldsymbol{y}\), where \(x_i\) denotes a word and \emph{L} denotes the length of the sentence. 
An IE model generates predictions \(\boldsymbol{\bar{y}}\).
The labeled training data \(\mathcal{D}_{\text {train }}^{S}=\{(\boldsymbol{x}, \boldsymbol{x^l}, \boldsymbol{y})\}\) is available for the source language, while only labeled testing data \(\mathcal{D}_{\text {test}}^{T}=\{(\boldsymbol{x}, \boldsymbol{x^l}, \boldsymbol{y})\}\) is available for the target language.
Formally, zero-shot cross-lingual IE aims at achieving good performance on \(\mathcal{D}_{\text {test}}^{T}\) by leveraging \(\mathcal{D}_{\text {train}}^{S}\).

\subsection{Basic Encoder}
\label{basicenc}
The basic encoder in this
paper consists of an mBERT~\cite{DBLP:conf/naacl/DevlinCLT19} and an \emph{L}-layer Transformer denoted as $\operatorname{Trans-L}$.
They are utilized to extract the contextual representations $\boldsymbol{h^{c}}$ and language-universal representations $\boldsymbol{h^{l}}$ respectively.

Following~\citet{DBLP:conf/aaai/AhmadPC21}, given a sentence \(\boldsymbol{x}\) of length \emph{L} from source language data \(\mathcal{D}_{\text {train}}^{S}\), three well-studied language-universal features are denoted as one-hot vectors \(\boldsymbol{x^p}\) for POS, \(\boldsymbol{x^d}\) for dependency relation, and \(\boldsymbol{x^e}\) for entity type.
Besides, the proposed constituent type vector \(\boldsymbol{x^c}\) is not a one-hot vector and its construction is described in Section~\ref{4.2.1}.
These calculations are formulated as:
\begin{align}
    \label{cat}
    \boldsymbol{x^{l}} &=[\boldsymbol{x^{p}} ; \boldsymbol{x^{d}} ; \boldsymbol{x^{e}} ; \boldsymbol{x^{c}}],  \\
    \boldsymbol{h^{c}} &= \operatorname{mBERT}(\boldsymbol{x}), \\
    \boldsymbol{h^{l}} &= \operatorname{Trans-L}(\boldsymbol{x^{l}}),
\end{align}
where $\boldsymbol{x^{l}}$ is the concatenation of four language-universal features.
\(\boldsymbol{h^{c}}=\{\boldsymbol {h}_i^{\boldsymbol c}\}_{i=1}^{L}\) and \(\boldsymbol{h}^{\boldsymbol l}=\{\boldsymbol {h}_i^{\boldsymbol l}\}_{i=1}^{L}\).
\(\boldsymbol h_i^{\boldsymbol c}\) and \(\boldsymbol h_{i}^{\boldsymbol l}\)  are representations of \(x_i\) and \(\boldsymbol x_i^{\boldsymbol l}\) respectively.

\subsection{Multi-level Interaction Network}\label{4.1}
Each type of language-universal features emphasizes its respective property during cross-language transfer, thus versatile abilities are exhibited when various features are available.
For example, POS can help augment the connection between words of the same part of speech across different languages, and dependency models the relationship between words to mitigate the word-order problem~\cite{DBLP:conf/emnlp/LiuCLZ19}. 
However, there is usually a semantic gap between these features and the contextual representations, leading to information loss since no explicit interactions are established between them when fusing these features.

In this section, a multi-level interaction framework is designed 
to hierarchically  interact the contextual representations \(\boldsymbol{h^{c}}\) and the language-universal representations \(\boldsymbol{h^{l}}\) at the global-, local- and task-level respectively, via an interactive loss mechanism. 
As  Figure~\ref{modelshine} depicts, 
the global-level interaction operates over the entire sentence to enhance the interaction between \(\boldsymbol{h^{c}}\) and \(\boldsymbol{h^{l}}\) as a whole.
Besides, the local-level divides a sentence into fixed-length sub-spans \(\boldsymbol{X^{s}}=\{\boldsymbol X_i^{\boldsymbol s}\}_{i=1}^{T}\) where \(\boldsymbol{X_i^{s}}=\{ x_j\}_{j=i}^{i+P-1}\), enhancing attention to span information.
Here \emph{P} is the length of each span and \emph{T} is the number of all sub-spans.
Lastly, the task-level interaction utilizes task-related mentions (such as entities, event triggers, etc.) to strengthen the model adaptation ability to different tasks.
In this way, various syntax features and contextual information can directly interact with each other at different levels, so that the representation capability can be enhanced at both word-level and span-level.
To measure the distribution discrepancy of two different random variables and effectively enable information sharing, a symmetrized Kullback–Leibler (KL) divergence~\cite{DBLP:conf/isit/Perez-Cruz08} is employed following~\citet{DBLP:conf/acl/JiangHCLGZ20} and ~\citet{DBLP:conf/acl/WangJBWHHT20} at each level of interaction.
$    {\rm KL}(P \| Q)=\sum\nolimits_{k} p_{k} \log \left(p_{k} / q_{k}\right)$ denotes the
KL-divergence of two discrete distributions $P$ and
$Q$ with the associated parameters of $p_{k}$ and $q_{k}$, respectively.
These calculations are formulated as:
\begin{align}
    \begin{small}
  \mathcal{L}_{g} = {\rm KL}(\boldsymbol{h^{c}}||\boldsymbol{h^{l}})+{\rm KL}(\boldsymbol{h^{l}}||\boldsymbol{h^{c}}),
    \end{small}
\end{align}
\vspace{-5mm}
\begin{small}
\begin{equation}
\begin{aligned}
    \label{local}
  \mathcal{L}_{l} = \frac{1}{T}\sum_{i=1}^{T}[ &{\rm KL}(\boldsymbol{H_i^{c}}||\boldsymbol{H_i^{l}}) + \\
                                            &{\rm KL}(\boldsymbol{H_i^{l}}||\boldsymbol{H_i^{c}}) ],
\end{aligned}
\end{equation}
\end{small}
\vspace{-5mm}
\begin{align}
    \begin{small}
  \mathcal{L}_{t} = {\rm KL}(\boldsymbol{H_{t}^{c}}||\boldsymbol{H_{t}^{l}})+{\rm KL}(\boldsymbol{H_{t}^{l}}||\boldsymbol{H_{t}^{c}}), 
    \end{small}
\end{align}
where \(\boldsymbol{H_i^{c}}=\{\boldsymbol h_j^{\boldsymbol c}\}_{j=i}^{i+P-1}\) and \(\boldsymbol{H_i^{l}}=\{\boldsymbol h_j^{\boldsymbol l}\}_{j=i}^{i+P-1}\). \(\boldsymbol{H_t^{c}}\) and \(\boldsymbol{H_t^{l}}\) are the task-related mention representations from mBERT
 and Transformer respectively.
 


\subsection{Constituency Structure Modeling}\label{4.2}
As aforementioned, previous studies focus
on encoding dependency structures, which describe only the dependency relationships between two words. 
However, the attributes of a constituent span in text and the relationships between multiple spans are crucial to many IE tasks. 
To this end, we introduce the constituency structure to construct span representation and to model the importance of each sub-span, to explicitly utilize span-level information to accurately find task-related spans for effective cross-language transfer.

\paragraph{Span Representation Construction}\label{4.2.1}
Since the constituent spans overlap each other, we first convert each span into a series of word-level one-hot vectors with BIO annotations (such as NP $\rightarrow$ B-NP, I-NP)~\cite{DBLP:conf/conll/Sang02}.
Then, these vectors are summed to derive $ \boldsymbol{x^c}\in  \mathbb{R}^{L\times C}$ mentioned in Section~\ref{basicenc}, where \emph{L} is the length of the sentence and \emph{C} is the number of the constituent types converted to BIO form. 
For example, given a sentence ``\emph{they have received deployment orders}'', we can derive the annotated constituent spans ``(0,0,NP), (1,4,VP), (2,4,VP), (3,4,NP), (0,4,S)'' with the Stanford
CoreNLP toolkit~\cite{DBLP:conf/acl/ManningSBFBM14}\footnote{(1,4,VP) denotes that a span constructed with 1st to 4th words in a sentence is a VP, i.e., verb phrase.}.
Then the ``\emph{deployment orders}'' is represented as ``B-NP I-NP''.
The details are shown in Table~\ref{example}.
The count-based representation can reflect the depth of words in the constituency tree to a certain extent and deeper words have larger counts.
In this way, the relationship between each word and each constituent type can be constructed to enrich representation information.

\begin{table}[t]
    \centering
    \setlength{\tabcolsep}{1.2mm}
    \begin{tabular}{l|cccccc}
    \toprule 
         \footnotesize{Words} &\footnotesize{B-NP} &\footnotesize{I-NP} &\footnotesize{B-VP} &\footnotesize{I-VP} &\footnotesize{B-S} &\footnotesize{I-S} \\\hline\hline
         they &1 &0 &0 &0 &1 &0 \\\hline
         have &0 &0 &1 &0 &0 &1 \\\hline
         received &0 &0 &1 &1 &0 &1 \\\hline
         deployment &1 &0 &0 &2 &0 &1 \\\hline
         orders &0 &1 &0 &2 &0 &1 \\\bottomrule
    \end{tabular}
    \caption{An example of the span representation construction for constituent spans in one sentence. Each element denotes the times each word occurs in all spans. 
    For instance, (1,4,VP) and (2,4,VP) both contain ``\emph{orders}'' in (orders, I-VP, 2), so the value is 2.
    }
\vspace{-3mm}
\label{example}
\end{table}

\paragraph{Sub-span Importance Construction}
Intuitively, a task-related phrase is more likely to be a constituent span or a sub-span of a constituent span. 
Despite the constituent type information for each word is modeled, the importance of each sub-span in the text also matters.
Those constituent spans or spans contained within constituent spans should be assigned higher importance to help find task-related phrases.
In this work, the importance of each sub-span in the text is modeled based on existing constituent spans.

\begin{table}[t]
    \centering
    \setlength{\tabcolsep}{3mm}
    \begin{tabular}{l|ccccc}
    \toprule 
         \large{Words} & \rotatebox{90}{\footnotesize{he}} & \rotatebox{90}{\footnotesize{bought}} & \rotatebox{90}{\footnotesize{apple}} & \rotatebox{90}{\footnotesize{juice}} & \rotatebox{90}{\footnotesize{here}} \\\hline\hline
         he &1 &1 &1 &1 &1 \\\hline
         bought &1 &1 &2 &2 &1 \\\hline
         apple &1 &2 &1 &3 &1 \\\hline
         juice &1 &2 &3 &1 &1 \\\hline
         here &1 &1 &1 &1 &1 \\\bottomrule
    \end{tabular}
    \caption{The frequency matrix shows the times of each sub-span in text is a sub-span of constituent spans. The annotated constituent spans are: (0,0,NP), (1,3,VP), (2,3,NP), (0,4,S).
    Considering that the attention between two words is bidirectional, we simply symmetrize the matrix.
    For the span (1,2), it is the sub-span of both (1,3) and (0,4), so the frequency is 2. Since a non-constituent word is usually not task-related (e.g. not an entity), we set it to 1.  }
\label{frequen}
\vspace{-3mm}
\end{table}

In detail, a frequency matrix denoted as $\boldsymbol{F}\in  \mathcal{R}^{L\times L}$ is constructed, where $F_{ij}$ represents the number of occurrences of span $(i,j)$ in all constituent spans. 
An example is illustrated in Table~\ref{frequen}.
A variant of Transformer $\operatorname{RTrans-N}(E,F)$ shown in Figure~\ref{modelshine} is designed by modifying the self-attention mechanism as: 
\begin{align}
\begin{small}
\label{revise}
\operatorname{Attn-F}(\boldsymbol{Q}, \boldsymbol{K}, \boldsymbol{V})=G(\operatorname{softmax}\left(\frac{\boldsymbol{Q} \boldsymbol{K}^{\boldsymbol{T}}}{\sqrt{d_{k}}}\right) \boldsymbol{V}).
\end{small}
\end{align}
Here, $\operatorname{softmax}$ function produces an attention matrix $\boldsymbol{A}\in  \mathcal{R}^{L\times L}$ where $A_{ij}$ denotes the attention weight that the $i$-th token pays to the $j$-th token in the sentence. 
$G$ is a function that 
integrates the frequency matrix to obtain modified attention weights.
The $(i,j)$-th element of the original attention matrix $A$ is modified as:
\begin{align}
G(A)_{i j}=\frac{F_{i j}A_{i j}}{\sum_{j} {F_{i j}}{A_{i j}}}.
\end{align}

Thus, the importance of each sub-span of a sentence is modeled according to the frequency it occurs across all constituent spans, with larger frequency values indicating higher importance. In this way, the model assigns more attention to sub-spans with higher importance.
Taking Table~\ref{frequen} as an example, the entity ``apple juice'' has the highest importance, so the model could find it accurately.
The final representation of the sentence $\boldsymbol{h^{f}}$ is formulated as:
\begin{align}
    \boldsymbol{h^{w}} &=\operatorname{Linear}([\boldsymbol{h^{c}}; \boldsymbol{h^{l}}]), \\
    \boldsymbol{h^{f}} &= \operatorname{RTrans-N}(\boldsymbol{h^{w},F}),
    \label{variant}
\end{align}
where $\operatorname{Linear}$ is a linear transformation.
$\boldsymbol{h^{w}}$ is the concatenation of contextual representations and language-universal representations.

\section{Downstream Tasks}
Three downstream tasks are employed to evaluate the effectiveness of the proposed SHINE encoder as comprehensively as possible, including the tasks of NER, relation extraction and event extraction. 

\subsection{Named Entity Recognition}
This task targets to locate and classify named entities in a text sequence.
Denote one sentence as \(\boldsymbol{x}=\{x_i\}_{i=1}^{L}\) with its labels \(\boldsymbol{y}=\{y_i\}_{i=1}^{L}\) and representation $\boldsymbol{h_{x}^{f}}$ from Eq.~(\ref{variant}), where \(y_i\) denotes the label of its corresponding word \(x_i\) and \emph{L} denotes the length of the sentence.
It is worth noting that this task aims at extracting entities, so the entity type $\boldsymbol{x^{e}}$ is removed from Eq.~(\ref{cat}). 
Following~\citet{DBLP:conf/acl/WuLKLH20},  $\boldsymbol{h_{x}^{f}}$ is fed into a softmax classification layer to calculate the probability of each word denoted as $ \boldsymbol{p}(x_{i})$ and cross-entropy loss is utilized to optimize the model:
\begin{align}
    &\mathcal{L}_{e}=\frac{1}{L}\sum_{i=1}^{L}\mathcal{L}_{\text {CE }}\left(\boldsymbol{p}(x_{i}), y_{i}\right).
\end{align}

Additionally, previous studies on NER also focus on distillation methods~\cite{DBLP:conf/acl/WuLKLH20,DBLP:conf/acl/ChenJW0G20,DBLP:conf/acl/LiHGCQZ22}.
In this work, the distillation method used by~\citet{DBLP:conf/acl/WuLKLH20} is adopted in this task for a fair and comprehensive comparison.
The source language model is denoted as the teacher model \(\Theta_{tea}\).
During distillation, a student model \(\Theta_{stu}\) with the same structure as \(\Theta_{tea}\) is distilled based on the unlabeled target language data \(\mathcal{D}_{\text {train}}^{T}\), which is fed into \(\Theta_{tea}\) to obtain its soft labels.
Given a sentence \(\boldsymbol{x^{\prime}}\) of length \emph{L}  from \(\mathcal{D}_{\text {train}}^{T}\),
the  objective for distillation~\cite{DBLP:journals/corr/HintonVD15} is to minimize the mean squared error (MSE) loss as:
\begin{align}
    \begin{small}
    \mathcal{L}_{\text {KD }}=\frac{1}{L}\sum_{i=1}^{L}{\text {MSE }}(\boldsymbol{p_{tea}}\left(x_{i}^{\prime}), \boldsymbol{p_{stu}}\left(x_{i}^{\prime} \right)\right).
    \end{small}
\end{align}
\subsection{Relation Extraction}
This task aims at predicting the relationship label of a pair of subject and object entities in a sentence.
Given two entity mentions $\boldsymbol{x_m}$ and
 $\boldsymbol{x_n}$ with representations $\boldsymbol{h_{m}^{f}}$ and $\boldsymbol{h_{n}^{f}}$ are derived respectively from Eq.~(\ref{variant}).
Following~\citet{DBLP:conf/aaai/AhmadPC21}, a sentence representation $\boldsymbol{h_{s}^{f}}$ is also obtained. 
Max-pooling is applied over these three vectors to derive the $\boldsymbol{\hat{h}_{m}^{f}}, \boldsymbol{\hat{h}_{n}^{f}}$, and  $\boldsymbol{\hat{h}_{s}^{f}}$.
Then the concatenation of the three vectors is fed into a softmax classification layer to predict the label as follows: 
\begin{align}
\begin{small}
\boldsymbol{p}_{mn}=\operatorname{softmax}\left(\boldsymbol{W}_{r}[\boldsymbol{\hat{h}_{m}^{f}};\boldsymbol{\hat{h}_{n}^{f}};\boldsymbol{\hat{h}_{s}^{f}}]+\boldsymbol{b}_{r}\right), 
\label{relation}
\end{small}
\end{align}
where  $\boldsymbol{W}_{r} \in R^{3 d_{\text {model }} \times r}$  and  $\boldsymbol{b}_{r} \in R^{r} $ are trainable parameters, and  $r$  is the total number of relation types. 
The objective of this task is to minimize the
cross-entropy loss as:
\begin{align}
    \mathcal{L}_{r}&=\sum_{m=1}^{M} \sum_{n=1}^{R_{i}}\mathcal{L}_{\text {CE }}\left( \boldsymbol{p}_{mn},y_{mn}\right), 
    \label{relation_loss}
\end{align}
where $M$ is the number of entity mentions, $R_{i}$ is the number of entity candidates for $i$-th entity mention and  $y_{mn}$ denotes the ground truth relation type between $\boldsymbol{x_m}$ and $\boldsymbol{x_n}$.

\subsection{Event Extraction}
This task can be decomposed into two sub-tasks of \textit{Event Detection} and \textit{Event Argument Role Labeling (EARL)}. 
Event detection aims at identifying event triggers and their types.
EARL predicts the argument candidate of an event trigger and assigns a role label to each argument from a pre-defined set of labels.
In this paper, we focus on EARL  and assume
event triggers of the input sentence are provided following~\citet{DBLP:conf/aaai/AhmadPC21}.

Given an event trigger $\boldsymbol{x_t}$ and an argument mention $\boldsymbol{x_a}$ with representation $\boldsymbol{h_{t}^{f}}$ and $\boldsymbol{h_{a}^{f}}$ respectively.
The concatenation of the three vectors
$\boldsymbol{\hat{h}_{t}^{f}}, \boldsymbol{\hat{h}_{a}^{f}}$, and  $\boldsymbol{\hat{h}_{s}^{f}}$ is fed into a softmax classification layer to calculate the probability of the argument role label $\boldsymbol{p}_{ta}$, following Eq.~(\ref{relation}).
The objective of this task $\mathcal{L}_{a}$ is to minimize the
cross-entropy loss following Eq.~(\ref{relation_loss}).
We change $M$ and $R_{i}$ with $N$ and $E_{i}$ respectively,
where $N$ is the number of event triggers, $E_{i}$ is the number of argument candidates for $i$-th event triggers.
$y_{mn}$ is changed with $y_{ta}$, denoting the ground truth argument role type between $\boldsymbol{x_t}$ and $\boldsymbol{x_a}$.

Finally, the loss for SHINE is as follows:
\begin{align}
\mathcal{L}_{f}=\mathcal{L}_{task}+\alpha(\mathcal{L}_{g}+\mathcal{L}_{l}+\mathcal{L}_{t}),
\label{finalloss}
\end{align}
where $\alpha$ is a manually set hyperparameter and 
$\mathcal{L}_{task}$ is the task loss $\mathcal{L}_{e}$, $\mathcal{L}_{r}$ or $\mathcal{L}_{a}$.

\section{Experiments}

\begin{table}[t]
\centering
\footnotesize
\setlength{\tabcolsep}{1mm}
\begin{tabular}{lcccc}
\toprule  
\textbf{Method} & \textbf{de} & \textbf{es} & \textbf{nl} & \textbf{Avg} \\ \midrule

BWET \cite{DBLP:conf/emnlp/XieYNSC18} &  57.76 & 72.37 & 71.25 & 67.13 \\ 
BS \cite{DBLP:conf/emnlp/WuD19} &  69.59 & 74.96 & 77.57 & 73.57 \\
TSL \cite{DBLP:conf/acl/WuLKLH20} &  73.16 & 76.75 & 80.44 & 76.78 \\ 
Unitrans \cite{DBLP:conf/ijcai/WuLKHL20} &  74.82 & 79.31 & 82.90 & 79.01  \\ 
AdvPicker \cite{DBLP:conf/acl/ChenJW0G20} &  75.01 & 79.00 & 82.90 & 78.97 \\ 
 \quad   w/o. KD &  73.89 & 76.92 & 80.62 & 77.14 \\ 
RIKD \cite{DBLP:conf/kdd/LiangGPSZZJ21} &  75.48 & 77.84 & 82.46 & 78.59 \\ 
\quad w/o. KD &  70.64 & 76.79 & 79.88 & 75.77 \\ 
TOF \cite{DBLP:conf/acl/ZhangMCXZ21} &  76.57 & 80.35 & 82.79 & 79.90 \\ 
MTMT \cite{DBLP:conf/acl/LiHGCQZ22} &  76.80 & \textbf{81.82} & 83.41 & 80.67 \\ 
\quad w/o. KD &  72.70 & 76.54 & 80.31 & 76.52 \\ 
\midrule
\textbf{SHINE} &  74.12 & 77.91 & 81.81 & 77.94 \\ 
\textbf{SHINE w. Distill}  & $\textbf{77.32}^{\dagger}$ & 80.86 & $\textbf{84.00}^{\dagger}$ & \textbf{80.73} \\ 

\midrule

SHINE w/o. interaction  &  73.68 & 77.42 & 81.01 & 77.37 \\ 
SHINE w/o. frequency &  73.75 & 77.52 & 81.23 & 77.50 \\ 
SHINE w/o. constituency &  73.39 & 76.78 & 80.55 & 76.91 \\ 
SHINE w/o. all &  73.13 & 76.39 & 80.38 & 76.63 \\ 
\bottomrule

\end{tabular}

    \caption{Evaluation results (\%) of entity-level F1-score on the test set of the CoNLL datasets~\cite{DBLP:conf/conll/Sang02,DBLP:conf/conll/SangM03}. Except MTMT w/o. KD was implemented by us,
    the other results were cited from the published literature.  For a fair comparison,
scores for the RIKD (mBERT) version were listed, and  scores for some distillation-based methods without distillation were also listed. Numbers marked with $\dagger$ denoted that the improvement over the best performing baseline was statistically significant (t-test with \(p\)-value \textless \(0.05\)).}
    \label{conll-result}
\vspace{-2mm}
\end{table}

\begin{table}[t]
\centering
\footnotesize
\setlength{\tabcolsep}{1.0mm}
\begin{tabular}{lcccc}
\toprule  
\textbf{Method} & \textbf{ar} & \textbf{hi} & \textbf{zh} & \textbf{Avg} \\ \midrule

 BS \cite{DBLP:conf/emnlp/WuD19} &  42.30 & 67.60 & 52.90 & 54.27 \\
 TSL \cite{DBLP:conf/acl/WuLKLH20} &  43.12 & 69.54 & 48.12 & 53.59 \\ 
 RIKD \cite{DBLP:conf/kdd/LiangGPSZZJ21} &  45.96 & 70.28 & 50.40 & 55.55 \\ 
 MTMT \cite{DBLP:conf/acl/LiHGCQZ22} &  52.77 & 70.76 & 52.26 & 58.60 \\ 
\midrule
\textbf{SHINE} &  51.61 & 68.67 & 53.47 & 57.91 \\ 
\textbf{SHINE w. Distill}  &  $\textbf{55.28}^{\dagger}$ & \textbf{71.10} & $\textbf{57.65}^{\dagger}$ & $\textbf{61.34}^{\dagger}$ \\ 
\midrule

SHINE w/o. interaction &  49.74 & 68.00 & 52.83 & 56.86 \\ 
SHINE w/o. frequency &  50.73 & 68.28 & 51.67 & 56.89 \\ 
SHINE w/o. constituency &  49.81 & 68.85 & 50.31 & 56.32 \\ 
SHINE w/o. all &  49.22 & 68.42 & 49.73 & 55.79 \\ 
\bottomrule

\end{tabular}

    \caption{Evaluation results (\%) of entity-level F1-score on the test set of the WikiAnn dataset~\cite{DBLP:conf/acl/PanZMNKJ17}. Results except ours were cited from the published literature.
    For a fair comparison,
scores of RIKD (mBERT) was listed.  Numbers marked with $\dagger$ denoted that the improvement over the best performing baseline was statistically significant (t-test with \(p\)-value \textless \(0.05\)).}
    \label{wiki-result}
\vspace{-3mm}
\end{table}

\subsection{Datasets}

We adopted CoNLL-2002~\cite{DBLP:conf/conll/Sang02}, CoNLL-2003~\cite{DBLP:conf/conll/SangM03} and WikiAnn~\cite{DBLP:conf/acl/PanZMNKJ17} datasets
for NER, 
Automatic Content Extraction
(ACE) 2005 dataset~\cite{DBLP:ace} for relation extraction and EARL.
Readers can refer to Appendix~\ref{sec-data} for the details of these datasets. 




For data preprocessing, UDPipe\footnote{https://lindat.mff.cuni.cz/services/udpipe/}~\cite{DBLP:conf/conll/StrakaS17} was applied for POS tagging and dependency parsing for CoNLL and WikiAnn datasets.
We followed~\citet{DBLP:conf/emnlp/SubburathinamLJ19} for the ACE 2005 datatset.
Finally, for all datasets, the Stanford
CoreNLP toolkit\footnote{https://stanfordnlp.github.io/CoreNLP/}~\cite{DBLP:conf/acl/ManningSBFBM14} was utilized for constituency parsing.

For NER, following previous work~\cite{DBLP:conf/acl/WuLKLH20},
English was employed as the source language and the other languages were
employed as the target languages.
Unlabeled
target language data in the training set and its language-universal features were utilized for distillation in NER.
As for relation extraction and EARL, all models were individually trained on one source
language, and directly evaluated on the other
two target languages following~\citet{DBLP:conf/aaai/AhmadPC21}.

\subsection{Evaluation Metrics}
For NER, an entity mention was correct if its offsets
and type matched a reference entity.
Following \citet{DBLP:conf/conll/Sang02}, entity-level F1-score was used as the evaluation metric.
The relation-level F1 score was employed for relation extraction~\cite{DBLP:conf/acl/LinJHW20,DBLP:conf/aaai/AhmadPC21}.
A relation mention was correct if its predicted type and the offsets of the two associated entity mentions were correct.
The argument-level F1 score was considered for EARL~\cite{DBLP:conf/emnlp/SubburathinamLJ19,DBLP:conf/aaai/AhmadPC21}. 
An event argument role label was correct if its
event type, offsets, and argument role type matched any
of the reference argument mentions.
Readers can refer to Appendix~\ref{sec-metrics} for the details of the metric calculations.

\subsection{Baselines and Implementation Details}
To compare SHINE on NER, 
these proposed approaches were chosen as baselines including:
(1) distillation-based methods: \textbf{TSL} \cite{DBLP:conf/acl/WuLKLH20}, \textbf{Unitrans} \cite{DBLP:conf/ijcai/WuLKHL20}, \textbf{AdvPicker} \cite{DBLP:conf/acl/ChenJW0G20}, \textbf{RIKD} \cite{DBLP:conf/kdd/LiangGPSZZJ21}, and \textbf{MTMT} \cite{DBLP:conf/acl/LiHGCQZ22}, and 
(2) non-distillation-based methods: \textbf{BWET}~\cite{DBLP:conf/emnlp/XieYNSC18}, 
\textbf{BS}~\cite{DBLP:conf/emnlp/WuD19} and \textbf{TOF}~\cite{DBLP:conf/acl/ZhangMCXZ21}.


As for relation extraction and EARL tasks, this method was mainly compared with the following: \textbf{CL-GCN}~\cite{DBLP:conf/emnlp/SubburathinamLJ19}, \textbf{CL-RNN}~\cite{DBLP:conf/emnlp/NiF19}, \textbf{OneIE}~\cite{DBLP:conf/acl/LinJHW20}, \textbf{GATE}~\cite{DBLP:conf/aaai/AhmadPC21}
and \textbf{CLEAE}~\cite{DBLP:conf/sigir/LouGYWZTX22}.
Readers can refer to Appendix~\ref{sec-baseline} and Appendix~\ref{sec-imple} for the implementation details of the baseline models and the proposed SHINE respectively.

\begin{table*}[t]
\centering
\footnotesize

\setlength{\tabcolsep}{1mm}
\resizebox{1.0\linewidth}{!}{

\begin{tabular}{l|ccccccc|ccccccc}
\toprule  

\multirow{6}{*}{\textbf{Method}}
&\multicolumn{7}{c}{\textbf{Relation Extraction}}
&\multicolumn{7}{|c}{\textbf{Event Argument Role Labeling}} \\
\cmidrule(r){2-15}

& $\begin{gathered}
\text{En} \\
\Downarrow \\
\text{Zh} \\
\end{gathered}$
& $\begin{gathered}
\text { En } \\
\Downarrow \\
\text { Ar } \\
\end{gathered}$
& $\begin{gathered}
\text { Zh } \\
\Downarrow \\
\text { En } \\
\end{gathered}$
& $\begin{gathered}
\text { Zh } \\
\Downarrow \\
\text { Ar } \\
\end{gathered}$
& $\begin{gathered}
\text { Ar } \\
\Downarrow \\
\text { En } \\
\end{gathered}$
& $\begin{gathered}
\text { Ar } \\
\Downarrow \\
\text { Zh } \\
\end{gathered}$
& \text{Avg} 

& $\begin{gathered}
\text {En} \\
\Downarrow \\
\text{Zh} \\
\end{gathered}$
& $\begin{gathered}
\text { En } \\
\Downarrow \\
\text { Ar } \\
\end{gathered}$
& $\begin{gathered}
\text { Zh } \\
\Downarrow \\
\text { En } \\
\end{gathered}$  
& $\begin{gathered}
\text { Zh } \\
\Downarrow \\
\text { Ar } \\
\end{gathered}$
& $\begin{gathered}
\text { Ar } \\
\Downarrow \\
\text { En } \\
\end{gathered}$
& $\begin{gathered}
\text { Ar } \\
\Downarrow \\
\text { Zh } \\
\end{gathered}$

& \text{Avg} \\ \midrule

 CL-GCN~\cite{DBLP:conf/emnlp/SubburathinamLJ19} & 41.16 & 42.74 & 46.60 & 39.90 & 39.12 & 36.45 & 40.99  &  38.89 & 39.70 & 36.58 & 36.99 & 35.57 &37.06  &  37.46  \\
 CL-RNN~\cite{DBLP:conf/emnlp/NiF19} & 47.01 & 45.60 & 48.18 & 39.80 & 40.31 & 42.26 & 43.86 &  44.97 & 39.05 & 40.69 & 40.16 & 35.34 & 37.22 &  39.57  \\ 
 OneIE~\cite{DBLP:conf/acl/LinJHW20} & 55.23 & 46.50 & 53.71 & 41.80 & 43.59 & 42.51 & 47.22 &  51.80 & 43.53 & 46.17 & 41.39 & 39.04 & 41.93 &  43.98  \\ 
 GATE~\cite{DBLP:conf/aaai/AhmadPC21} & 53.52 & \textbf{50.77} & 52.25 & 45.36 & 41.67 & 44.14 & 47.95 &  48.61 & 50.18 & 45.99 & 45.04 & 42.52 & 38.39 & 45.12   \\ 
 CLEAE~\cite{DBLP:conf/sigir/LouGYWZTX22} & 54.63 & 46.91 & 55.87 & 44.52 & 42.23 & 46.03 & 48.36 
 &  53.96 & 51.12 & 47.83 & 45.91 & \textbf{45.14} & 43.36 & 47.89   \\ 
 
\midrule
\textbf{SHINE} & $\textbf{59.39}^{\dagger}$ & 47.62 & $\textbf{56.41}^{\dagger}$ & $\textbf{46.66}^{\dagger}$ & $\textbf{44.56}^{\dagger}$ & $\textbf{47.22}^{\dagger}$ & $\textbf{50.31}^{\dagger}$
&  $\textbf{57.92}^{\dagger}$ & \textbf{51.60} & $\textbf{51.35}^{\dagger}$ & $\textbf{48.11}^{\dagger}$ & 41.86  & $\textbf{45.49}^{\dagger}$ &  $\textbf{49.38}^{\dagger}$ \\ 

\midrule
SHINE w/o. interaction &  58.26 & 47.03 & 56.30 & 45.15 & 42.89 & 46.07 & 49.28 
& 56.86 &  51.01 & 51.30 & 46.94 & 41.10 & 42.04 & 48.21  \\ 
SHINE w/o. frequency & 58.87 & 46.96 & 56.23 & 45.31 & 43.36 & 45.68 & 49.40
&  56.99 & 50.90 & 50.92 & 47.31 & 40.79 & 43.67 &  48.43  \\ 
SHINE w/o. constituency & 56.79 & 46.52 & 55.56 & 44.64 & 42.01 & 45.58 & 48.51
&  56.60 & 49.77 & 50.65 & 46.42 & 39.37 & 40.81 &  47.27 \\ 
SHINE w/o. all & 56.05 & 45.98 & 55.03 & 42.83 & 40.76 & 44.63 & 47.54
&  56.12 & 48.18 & 50.54 & 44.67 & 38.29 & 37.18 &  45.83  \\

\bottomrule

\end{tabular}

}
    \caption{Evaluation results (\%) of F1-score on the test set of the ACE 2005 dataset~\cite{DBLP:ace} for relation extraction and EARL. Results except ours were obtained by implementing the source code of the baseline models provided by the authors. Languages
on top and bottom of $\Downarrow$ denoted the source and target languages respectively.
Numbers marked with $\dagger$ denoted that the improvement over the best performing baseline was statistically significant (t-test with \(p\)-value \textless \(0.05\)).}
    \label{ace-result}

\end{table*}

\subsection{Results and Comparison}
Table~\ref{conll-result}, \ref{wiki-result} and \ref{ace-result} reported the zero-shot cross-lingual IE results of different methods on 3 tasks and 4 benchmarks, containing 7 target languages.
For the NER task, the results show that the proposed \textbf{SHINE w. Distill} method outperformed MTMT (previous SOTA) on average by absolute margins of 0.06\% and 2.74\% in terms of CoNLL and WikiAnn datasets respectively.
Besides, the proposed SHINE also significantly outperformed baseline distillation method TSL on average by absolute margins of 1.16\% and 4.32\% in terms of CoNLL and WikiAnn datasets respectively, demonstrating the effectiveness of these language-universal features.
For a fair comparison, we compared SHINE against the version of Advpicker w/o. KD, RIKD w/o. KD, MTMT w/o. KD, and the results showed that SHINE significantly outperformed them.
Our results also demonstrated
 the compatibility and scalability between SHINE and distillation.

As for relation extraction and EARL, SHINE outperformed 
CLEAE (previous SOTA) in most all transfer directions 
with an average improvement of 1.95\% and 1.49\% in relation extraction and EARL tasks respectively.
It was worth noting that GATE used mBERT as the contextual representation extractor without fine-tuning.
This might lead to the cross-lingual information in mBERT not being fully exploited, subsequently degrading the performance of the model.
Our results clearly demonstrate that SHINE is highly effective and generalizes well across languages and tasks.

\subsection{Analysis}

\vspace{-1mm}
\paragraph{Ablation Study}
To validate the contributions of different components in SHINE, the following variants and baselines were conducted to perform the ablation study:
 (1) SHINE w/o. interaction, which removed the hierarchical interaction framework.
Besides, the constituency feature was still used during training.
 (2) SHINE w/o. frequency, which removed the frequency matrix.
Besides, the constituent span embeddings constructed in Section~\ref{4.2.1} and 
 the interaction framework were still used during training.
 (3) SHINE w/o. constituency, which removed constituent span embeddings and frequency matrix.
Besides, the interaction framework was still utilized.
 (4) SHINE w/o. all, which removed all the components mentioned above. 
It was the base structure of the proposed SHINE. 

Results of the ablation experiments were shown
in the bottom four lines of Table~\ref{conll-result}, \ref{wiki-result} and \ref{ace-result} respectively. 
 Additional in-depth analyses reveal:
 (1) removal of the interaction network (SHINE vs SHINE w/o. interaction) caused a significant performance drop, demonstrating the importance of establishing interaction between language-universal features and contextual information, and (2) removing constituency features caused significant performance drops (SHINE vs SHINE w/o. frequency and SHINE w/o. constituency). Both constituent type embeddings and sub-span importance information were useful.

The ablation study validated the effectiveness of all components.
Moreover, the subtle integration of these modules achieved highly promising performance.
Not only hierarchical interaction should be established to capture the complementary
 information of features and contextual information, but also constituency structure should be modeled for effective cross-lingual transfer.

 \vspace{-1mm}
 \paragraph{Case Study}
 To further illustrate the  effectiveness of SHINE and to explore how language-universal features play a role in cross-language transfer, the embedding distribution of three models was shown in Appendix~\ref{sec-case}.
 Distribution discrepancy within SHINE was significantly smaller than the base model.
 It shows that with the guidance of language-universal features,
 the proposed encoder could capture complementary information to alleviate discrepancy between languages to
 derive language-agnostic contextualized representations.

\section{Conclusion}
In this paper, we propose a syntax-augmented hierarchical interactive encoder for zero-shot cross-lingual IE. 
A multi-level interaction framework is designed to interact the complementary structured information contained in language-universal features and contextual representations.
Besides, the constituency structure is first introduced to explicitly model and utilize the constituent span information.
Experiments show that the proposed method achieves highly promising performance across seven languages on three tasks and four benchmarks.
In the future, we will extend this method to more languages and tasks.
Besides, we will explore other sources of language-universal features as well as the relationships between features 
to augment representation capability.


\section*{Limitations}
Although the proposed method has shown great performance for zero-shot cross-lingual IE,
we should realize that the proposed method still can be further improved.
For example, relationships between different language-universal features should be considered because there is an implicit mutual influence between them,  explicitly modeling it can make these features organically form a whole.
Then the model can use this information more efficiently and it will be a part of our future work.
In addition, some languages cannot be supported by existing language tools.
Although toolkits for some similar languages can be utilized for annotating, denoising these annotations is worth studying.

\bibliography{main}
\bibliographystyle{acl_natbib}

\clearpage
\appendix
\section{Appendices}

\subsection{Datasets}
\label{sec-data}

We adopted 
CoNLL\footnote{http://www.cnts.ua.ac.be/conll2003}~\cite{DBLP:conf/conll/Sang02,DBLP:conf/conll/SangM03} and WikiAnn\footnote{http://nlp.cs.rpi.edu/wikiann}~\cite{DBLP:conf/acl/PanZMNKJ17} datasets
for NER,
Automatic Content Extraction
(ACE) 2005 dataset\footnote{https://catalog.ldc.upenn.edu/LDC2006T06}~\cite{DBLP:ace} for relation extraction and EARL.
\paragraph{CoNLL}included two datasets:
(1) CoNLL-2002 (Spanish, Dutch);
(2) CoNLL-2003 (English, German);
They were annotated with 4 entity types:
LOC, MISC, ORG, and PER.

\paragraph{WikiAnn} included
English, Arabic, Hindi, and Chinese.
It was annotated with 3 entity types: LOC, ORG, and PER.
CoNLL and WikiAnn datasets were annotated with the BIO entity
labelling scheme and were divided into the training,
development and testing sets. Table~\ref{ner} shows the
statistics of these datasets.

\paragraph{ACE 2005}  included three languages (English, Chinese and Arabic).
It defined an ontology that included 7 entity types, 18 relation subtypes, and 33 event subtypes.
We added a class label \emph{None} to denote that two entity mentions or a pair of an event mention and an argument candidate under consideration did not have a relationship belonging to the target ontology.
Since the official did not divide the training, development and testing sets,
We adopted the same dataset-splitting strategy following~\citet{DBLP:conf/emnlp/SubburathinamLJ19}.
Besides, we re-implemented these baseline models based on the code provided by authors using default settings, which were described in Appendix~\ref{sec-baseline} . 
Table~\ref{ace} shows the
statistics of the dataset.

\begin{table}[t]
\centering
\footnotesize
\setlength{\tabcolsep}{1.8mm}{
\begin{tabular}{ccccc}
\toprule  
Language & Type & Train & Dev & Test \\ \midrule 
\multicolumn{5}{c}{CoNLL dataset~\cite{DBLP:conf/conll/Sang02,DBLP:conf/conll/SangM03}} \\ 
\midrule 
English-en &  Sentence & 14,987 & 3,466 & 3,684 \\
(CoNLL-2003) & Entity & 23,499 & 5,942 & 5,648 \\
\hline German-de & Sentence & 12,705 & 3,068 & 3,160 \\
(CoNLL-2003) & Entity  & 11,851 & 4,833 & 3,673 \\
\hline Spanish-es & Sentence  & 8,323 & 1,915 & 1,517 \\
(CoNLL-2002) & Entity  & 18,798 & 4,351 & 3,558 \\
\hline Dutch-nl & Sentence  & 15,806 & 2,895 & 5,195 \\
(CoNLL-2002) & Entity  & 13,344 & 2,616 & 3,941 \\
\midrule
\multicolumn{5}{c}{WikiAnn dataset~\cite{DBLP:conf/acl/PanZMNKJ17}} \\ 
\midrule 
\multirow{2}{*}{English-en} &  Sentence & 20,000 & 10,000 & 10,000 \\ 
 & Entity & 27,931 & 14,146 & 13,958 \\
\hline \multirow{2}{*}{Arabic-ar} & Sentence & 20,000 & 10,000 & 10,000 \\
 & Entity  & 22,500 & 11,266 & 11,259 \\
\hline \multirow{2}{*}{Hindi-hi} & Sentence  & 5,000 & 1,000 & 1,000 \\
 & Entity  & 6,124 & 1,226 & 1,228 \\
\hline \multirow{2}{*}{Chinese-zh} & Sentence  & 20,000 & 10,000 & 10,000 \\
 & Entity  & 25,031 & 12,493 & 12,532 \\
\bottomrule
\end{tabular}}
    \caption{The statistics of the CoNLL~\cite{DBLP:conf/conll/Sang02,DBLP:conf/conll/SangM03} and WikiAnn~\cite{DBLP:conf/acl/PanZMNKJ17} datasets.}
    \label{ner}
\end{table}

\begin{table}[t]
\centering
\footnotesize
\setlength{\tabcolsep}{2.4mm}{
\begin{tabular}{ccccc}
\toprule  
Language & Type & Train & Dev & Test \\ \midrule

\multirow{4}{*}{English-en} &  Sentence & 14,671 & 873 & 711 \\
 & Event & 4,317 & 492 & 422 \\
  & Argument & 7,814 & 933 & 892 \\
   & Relation & 5,247 & 550 & 509 \\
\hline \multirow{4}{*}{Chinese-zh} & Sentence & 5,847 & 715 & 733 \\
 & Event  & 2,610 & 325 & 336 \\
  & Argument & 6,281 & 727 & 793 \\
   & Relation & 5,392 & 807 & 664 \\
\hline \multirow{4}{*}{Arabic-ar} & Sentence  & 2,367 & 203 & 210 \\
 & Event & 921 & 141 & 127 \\
  & Argument & 2,523 & 328 & 367 \\
   & Relation & 2,389 & 266 & 268 \\

\bottomrule

\end{tabular}}
    \caption{The statistics of the ACE 2005 dataset~\cite{DBLP:ace}.}
    \label{ace}
\end{table}

\subsection{Evaluation Metrics}
\label{sec-metrics}
For NER, following \citet{DBLP:conf/conll/Sang02}, entity-level F1-score was used as the evaluation metric.
Denote \emph{A} as the number of all entities classified by the model, \emph{B} as the number of all correct entities classified by the model, and \emph{E} as the number of all correct entities, the precision (P), recall (R), and entity-level F1-score (E-F1) of the model were:
\begin{align}
    \operatorname{P} = \frac{B}{A}, \ \ \operatorname{R} = \frac{B}{E}, \ \ \operatorname{E-F1} = \frac{2 \times P \times R}{P+R}.
\end{align}

Following previous works~\cite{DBLP:conf/acl/LinJHW20,DBLP:conf/aaai/AhmadPC21}, the relation-level F1 score was considered for relation extraction. 
A relation mention was correct if its predicted type and
the offsets of the two associated entity mentions are correct.
Denote \emph{A} as the number of all relation mentions classified by the model, \emph{B} as the number of all correct relation mentions  classified by the model, and \emph{E} as the number of all correct relation mentions, the precision (P), recall (R), and relation-level F1-score (R-F1) of the model were:

\begin{align}
    \operatorname{P} = \frac{B}{A}, \ \ \operatorname{R} = \frac{B}{E}, \ \ \operatorname{R-F1} = \frac{2 \times P \times R}{P+R}.
\end{align}

An event argument role label was correct if its
event type, offsets, and argument role type matched any
of the reference argument mentions~\cite{DBLP:conf/aaai/AhmadPC21}.
Denote \emph{A} as the number of all arguments classified by the model, \emph{B} as the number of all correct arguments classified by the model, and \emph{E} as the number of all correct arguments,
the argument-level F1 score (A-F1) was defined as similar to relation-level F1 score:
\begin{align}
    \operatorname{P} = \frac{B}{A}, \ \ \operatorname{R} = \frac{B}{E}, \ \ \operatorname{A-F1} = \frac{2 \times P \times R}{P+R}.
\end{align}




\subsection{Baseline Models} 
\label{sec-baseline}
We described the implementation details for all the
models for NER as follows:

\noindent
\textbf{TSL} \cite{DBLP:conf/acl/WuLKLH20} proposed a teacher-student
learning model, via using source-language models as teachers to train a student model on unlabeled data in the target language for cross-lingual NER. The number of parameters of this model was about 110M.

\noindent
\textbf{Unitrans} \cite{DBLP:conf/ijcai/WuLKHL20}  unified both model transfer and data transfer based on their complementarity via enhanced knowledge distillation on unlabeled target-language data. The number of parameters of this model was about 331M.

\noindent
\textbf{AdvPicker} \cite{DBLP:conf/acl/ChenJW0G20} proposed a novel approach to combine the feature-based method and pseudo labeling via language adversarial learning for cross-lingual NER. The number of parameters of this model was about 178M.

\noindent
\textbf{RIKD} \cite{DBLP:conf/kdd/LiangGPSZZJ21} proposed a reinforced knowledge distillation framework. The number of parameters of this model was about 111M.

\noindent
\textbf{MTMT} \cite{DBLP:conf/acl/LiHGCQZ22} proposed an unsupervised multiple-task and multiple-teacher model for cross-lingual NER. The number of parameters of this model was about 220M.

In addition, \textbf{BWET}~\cite{DBLP:conf/emnlp/XieYNSC18}, 
\textbf{BS}~\cite{DBLP:conf/emnlp/WuD19} and \textbf{TOF}~\cite{DBLP:conf/acl/ZhangMCXZ21} were non-distillation-based methods.
The number of parameters of these model was about 12M, 111M and 114M respectively.

Furthermore,  baselines for relation extraction and EARL are shown below:

\noindent
\textbf{CL-GCN}
\cite{DBLP:conf/emnlp/SubburathinamLJ19}
used GCN~\cite{DBLP:conf/iclr/KipfW17} to embed the language-universal feature and context to learn structured  space representation.
We refer to the released code from \citet{DBLP:conf/aaai/AhmadPC21}\footnote{https://github.com/wasiahmad/GATE}
to re-implement it. 

\noindent
\textbf{CL-RNN}
\cite{DBLP:conf/emnlp/NiF19}
utilized BiLSTM to embed the language-universal feature and  contextual representations to learn structured  space representation.
We refer to the released code from \citet{DBLP:conf/aaai/AhmadPC21}
to re-implement this method. 

\noindent
\textbf{OneIE}
\cite{DBLP:conf/acl/LinJHW20}
was a joint model trained with multitasking to capture cross-subtask and cross-instance
inter-dependencies. We used their provided
code\footnote{http://blender.cs.illinois.edu/software/oneie/} to train the model with default settings. The number of parameters of this model was about 115M.

\noindent
\textbf{GATE}
\cite{DBLP:conf/aaai/AhmadPC21}
proposed a Transformer layer with syntactic distance to encode the language-universal features. 
The number of parameters of this model was about 118M.
We utilized the official release code to
re-implement this method. 

\noindent
\textbf{CLEAE}
\cite{DBLP:conf/sigir/LouGYWZTX22}
proposed a translation-based method with an implicit annotation. 
We extend this method to relation extraction as it is similar to EARL.
The number of parameters of this model was about 112M.
The official release code\footnote{https://github.com/HLT-HITSZ/CLEAE} was utilized to re-implement it.








\subsection{Implementation Details} 
\label{sec-imple}
All code was implemented in the PyTorch framework\footnote{https://pytorch.org/}.
All of the context encoders mentioned in this paper employed pre-trained cased mBERT \cite{DBLP:conf/naacl/DevlinCLT19} in HuggingFace’s Transformers\footnote{https://huggingface.co/transformers} where the number of transformer blocks was 12, the hidden layer size was 768, and the number of self-attention heads was 12.
Following \citet{DBLP:conf/emnlp/KeungLB19}, some hyperparameters were tuned on each target language.
All language-universal feature encoders mentioned in this paper employed 2-layer Transformer~\cite{DBLP:conf/nips/VaswaniSPUJGKP17}, the hidden layer size was 768, and the number of self-attention heads was 8.
For the variant transformer mentioned in Eq.~(\ref{variant}), the layer was set to 1, the hidden size was 768, and the number of self-attention heads was 8.
The $\alpha$ in Eq.~(\ref{finalloss}) was set to 10 and the span length \emph{P} utilized in Eq.~(\ref{local}) was set to 4.

For NER, following~\citet{DBLP:conf/acl/WuLKLH20}, each batch contained 32 examples, with a maximum encoding length of 128.
The dropout rate was set to 0.1, and AdamW \cite{DBLP:conf/iclr/LoshchilovH19} with WarmupLinearSchedule in the Transformers Library was used as optimizer.
The learning rate was set to 5e-5 for the teacher models and  2e-5 for the student models.
Following~\cite{DBLP:conf/emnlp/WuD19}, the parameters of the embedding layer and the bottom three layers of the mBERT used in the teacher model and the student model were frozen.
All models were trained for 10 epochs and chosen the best checkpoint with the target dev set.
For relation extraction and EARL, the other parameters were set following~\citet{DBLP:conf/acl/LinJHW20}.
We optimized our model with Adam~\cite{DBLP:journals/corr/KingmaB14} for 80
epochs with a learning rate of 5e-5 and a dropout rate of 0.1.
Furthermore, each experiment was conducted 5 times and reported the mean F1-score.

The number of parameters of a model was about 130M.
The whole training of SHINE was implemented with one GeForce RTX 3090, which consumed about 3 hours for NER and 8 hours for relation extraction and EARL.

\subsection{Case Study} \label{sec-case}

\begin{figure}[h]
\centering
\includegraphics[width=0.45\textwidth]{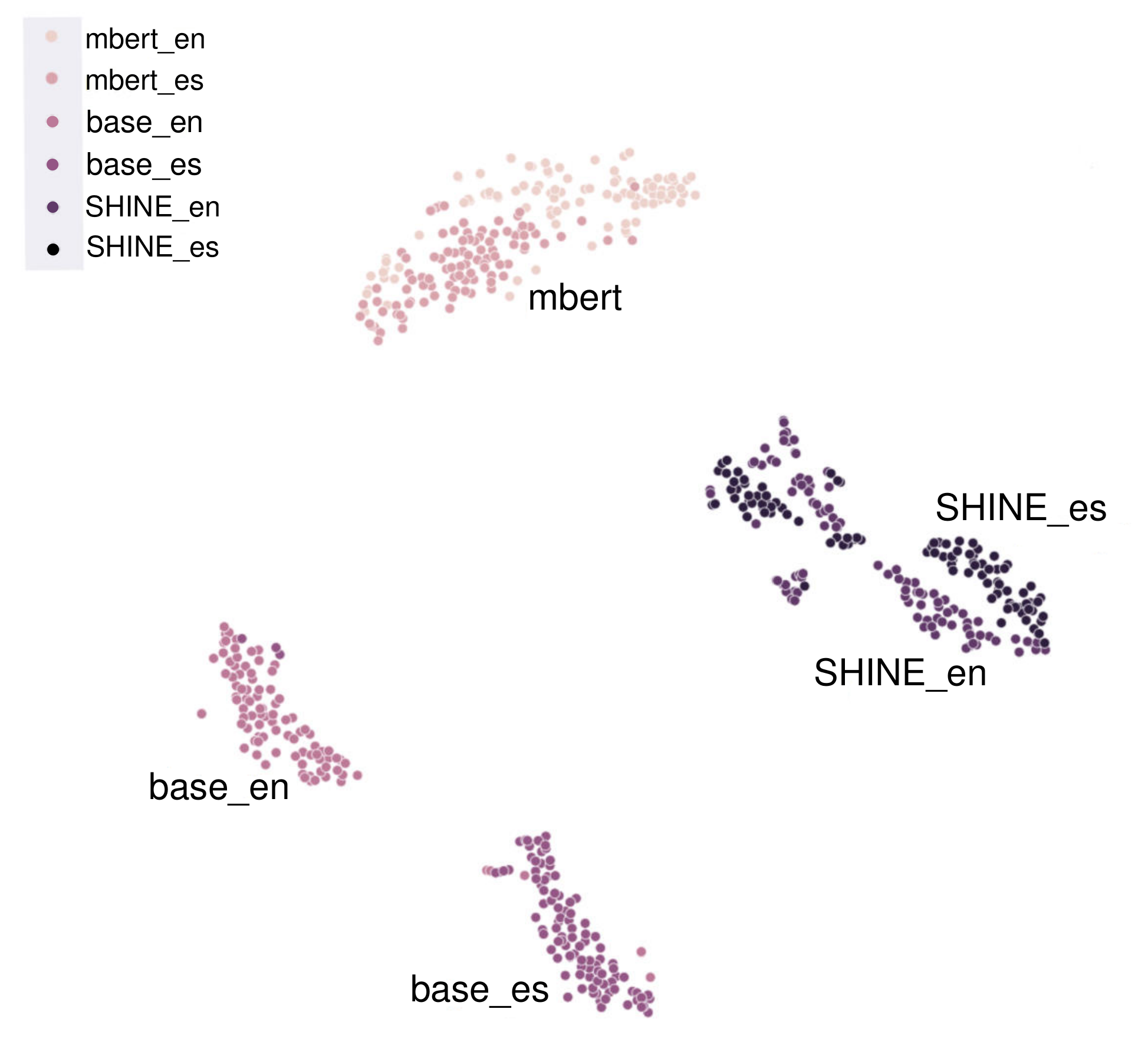}
\caption{
T-SNE visualization~\cite{van2008visualizing} of semantic domains of different models by randomly sampling 100 unannotated English (en, source) and Spanish (es, target) sentences from the training set of the CoNLL datasets~\cite{DBLP:conf/conll/Sang02,DBLP:conf/conll/SangM03}.
``\emph{mbert}'' refers to the untrained mBERT and
``\emph{base}'' refers to the ``\emph{BERT-Softmax}'' model without language-universal features, which is the backbone of ~\citet{DBLP:conf/acl/WuLKLH20} and ~\citet{DBLP:conf/acl/LiHGCQZ22}. 
``\emph{SHINE}'' refers to the model proposed in this paper trained on English language of the CoNLL dataset.
Each point refers to the average token representation of a sample in source/target languages. }   
\label{case}
\end{figure}

Figure~\ref{case} shows the representation distribution for the two languages across the three models.

\end{document}